\documentclass[preprint,12pt,authoryear]{elsarticle}
\usepackage{amssymb}
\journal{Knowledge Base Systems}

\usepackage{times}
\usepackage{latexsym}
\usepackage[utf8]{inputenc}
\usepackage{url}
\usepackage{paralist}
\usepackage{amsmath}
\usepackage{comment}
\usepackage{listings}
\usepackage{color}
\usepackage[table]{xcolor}
\usepackage{pdflscape}
\usepackage{graphicx}
\begin{document}

\begin{frontmatter}

\title{Why do you say they are similar? \\ Interpretable Semantic Textual Similarity}

\author{I.~Lopez-Gazpio}
\author{M.~Maritxalar}
\author{A.~Gonzalez-Agirre}
\author{G.~Rigau}
\author{L.~Uria}
\author{E.~Agirre\corref{cor1}}
\ead{e.agirre@ehu.eus}

\cortext[cor1]{Corresponding author}
\address{IXA NLP group, Informatics Faculty UPV/EHU, Manuel Lardizabal 1, 20008 - Donostia, Basque Country}

\begin{abstract}
  User acceptance of artificial intelligence agents might depend on
  their ability to explain their reasoning, which requires adding an
  interpretability layer that facilitates users to understand their
  behavior. This paper focuses on adding an interpretable layer on top
  of Semantic Textual Similarity (STS), which measures the degree of
  semantic equivalence between two sentences. The interpretability
  layer is formalized as the alignment between pairs of segments
  across the two sentences, where the relation between the segments
  is labeled with a relation type and a similarity score.  We present
  a publicly available dataset of sentence pairs annotated following
  the formalization.  We then develop a system trained on this dataset
  which, given a sentence pair, explains what is similar and different,
  in the form of graded and typed segment alignments. When evaluated on the dataset, the system
  performs better than an informed baseline, showing that the dataset
  and task are well-defined and feasible. Most importantly, two user
  studies show
  how the system output can be used to automatically 
  produce explanations  in natural language. Users 
  performed better when having access to the explanations, providing
  preliminary evidence that our dataset and method to automatically produce
  explanations is  useful in real applications.
\end{abstract}


\end{frontmatter}

\section{Introduction}
\label{Introduction}

Since the early days of expert systems, it is acknowledged that one key factor for users and domain experts to accept expert systems in real-world domains is the ability of the expert systems to explain their reasoning \citep[p. 336]{buchanan1984rule,KER:146473,Korb:2010:BAI:1941985}. We also think that user acceptance of artificial intelligence agents will depend on their ability to explain their reasoning, which requires adding an interpretability layer to facilitate users to understand their behavior.

Our work explores interpretability in the context of Semantic Textual Similarity (STS) \citep{agirre-EtAl:2012:STARSEM-SEMEVAL}. STS measures semantic equivalence between two text snippets using graded similarity, capturing the notion that some sentences are more similar than others, ranging from   
complete unrelatedness up to semantic equivalence. Systems attaining high correlations with gold truth scores have been routinely reported \citep{agirre-EtAl:2015:SemEval}.
As an example of the STS task, given the following two sentences drawn from a corpus of News headlines, the annotators judged its similarity as ``roughly equivalent, but some minor information differs'':
\begin{quote}
    12 killed in bus accident in Pakistan\\
    10 killed in road accident in NW Pakistan
\end{quote}

Our final goal is to build systems that are able to explain which are the differences and commonalities between any two sentences. The output for the two sample sentences would be something like the following: 

\begin{quote}
The two sentences talk about accidents with casualties in Pakistan, but they differ in the number of people killed (12 vs. 10) and level of detail: the first one specifies that it is a \textit{bus} accident, and the second one specifies that the location is \textit{NW} Pakistan. 
\end{quote}

While giving such explanations comes naturally to people, constructing algorithms and computational models that mimic human level performance represents a difficult natural language understanding problem. In this article we define a first step of such an ambitious goal.
We build and evaluate a system that, given two sentences, returns a textual explanation of the commonalities and differences between the two sentences. The system is based on a formalization of the interpretability layer as an explicit alignment of segments in the two sentences, where alignments are annotated with a relation type and a similarity score. The core part of the system is trained and evaluated on a dataset of sentence pairs which has been annotated with the alignments. The trained system is thus able to return the reasons for the similarity between the two sentences in the form of typed segment alignments. The evaluation on the annotated dataset shows that the system performs better than an informed baseline, showing that the task is well-defined and feasible.

\begin{figure*}[t]
 \begin{center}
   \includegraphics[scale=0.75]{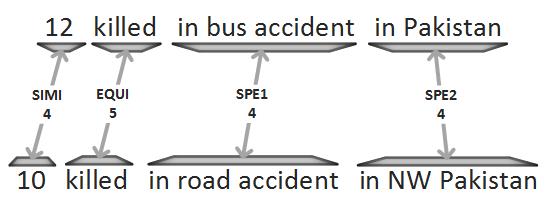}
 \end{center}
  \caption{Graphical representation of the interpretability layer. 
The two sentences were split in four segments each, which were aligned as follows: \emph{``12'} is similar to \emph{``10''} with a similarity score of 4 (SIMI 4), \emph{``killed''} is equivalent to \emph{``killed''} with score 5 (EQUI 5), \emph{``in bus accident''} is more specific than \emph{``in road accident''} with score 4 (SPE1 4), and \emph{``in Pakistan''} is more general than \emph{``in NW Pakistan''} with score 4 (SPE2 4). See Section \ref{sec:annotation-procedure} for more details on the annotation procedure.}
  \label{example}
\end{figure*}

Figure \ref{example} shows the formalization of the interpretability layer for the two sample sentences, including segments, alignments, types and scores of the alignments. Types include relations like equivalence, opposition, specialization, similarity or relatedness. The similarity scores for aligned segments range from 0 (no relation) to 5 (equivalence). 

In addition to the dataset and core system, we also build a verbalization system, that is, a system which takes as input the alignments and produces a human-readable explanation based on templates. This system returns the following text for the alignment in Figure \ref{example}:
\begin{quote}
The two sentences are very similar.
Note that 'in bus accident' is a bit more specific than 'in road accident' in this context.
Note also that '12' and '10' are very similar in this context.
Note also that 'in Pakistan' is a bit more general than 'in NW Pakistan' in this context.
\end{quote}

In order to measure the quality and usefulness of the explanations, direct comparison to human-elicited text (e.g. the explanation above) is problematic, and would not tell us about the usefulness. Instead, we measure whether the automatically produced explanations are useful in two user studies. In the first study, English native speakers scored the similarity of sentence pairs, with and without automatically produced explanations. In the second study, we simulated a tutoring scenario where students were graded with respect to a reference sentence. The users, simulating to be students, had to state whether they agreed with the grade, with and without access to the automatically produced explanations. Both studies show that users that read the explanations agreed with the system scores more often than users which did not have access to explanations. 

We summarize the contributions of this article as follows:
\begin{itemize}
\item It formalizes the interpretability layer in the context of STS as a graded and typed alignment between segments in the two sentences.
\item It describes a publicly available  dataset of sentence pairs (coming from news headlines and image captions) annotated with the interpretability layer following the above formalization.
\item It describes a system that, given two sentence pairs, is able to return alignments between the segments in the two sentences annotated with relation type and a graded similarity score. The system is trained and evaluated in the annotated dataset, with good results, well above an informed baseline and in the state-of-the-art.

\item It presents an extension of the system which returns a textual explanation of the reasons for the similarity judgment.
\item It shows two user-studies where the automatically produced explanations help users to better attain their tasks, providing preliminary evidence that our formalization and specific system are  useful in real applications. 
\end{itemize}

This paper is organized as follows. We first introduce related work. Section \ref{TaskDescription} deals with the creation of the dataset, followed by Section \ref{formalization}, which presents a comparison to a related dataset. Section \ref{SystemConstruction} explains the system, and Section  \ref{Evaluation} its evaluation. Section \ref{ApplicationInterpretable} presents the user study, and, finally, Section \ref{ConclusionsFutureWork} draws the conclusions.


\section{Related work}
\label{RelatedWork}

Early work on adding explanations in the context of bayesian networks includes both visualizations and verbal explanations about the model itself or the conclusions drawn about the domain \citep{opac-b1001136,Suermondt:1992:EBB:143629}. For instance, Elvira \citep{consortium2002elvira}
is a Bayesian Network package that offers both verbal explanations (about nodes and arcs) as well
as graphical explanations. 

Explanations are important in the teaching domain, where Intelligent Tutoring Systems (ITS) strive to provide feedback beyond correct/incorrect judgments. In most cases the systems rely on costly domain-dependent and question-dependent knowledge \citep{Aleven01pedagogicalcontent,jordan2006natural}, but some scalable alternatives based on generic Natural Language Processing (NLP) techniques are also available \citep{nielsen2009recognizing}. Our approach is related in spirit with this last paper, but we formalize the interpretability layer differently, as we will see below. 

In the area of NLP, the interpretability of representation models learned from raw data is also a widespread concern. \citet{ritter-mausam-etzioni:2010:ACL} show that they are able to infer classes which are easily interpretable by humans, and \citet{fyshe-EtAl:2015:NAACL-HLT} argue that the dimensions of their word representations correspond to easily interpretable concepts. To our knowledge this article is the first research work in the area of NLP addressing explicit human-readable explanations. 

Our work is situated in the area of Natural Language Understanding, where two related enabling tasks have been extensively used to evaluate the quality of the semantic representations, Semantic Textual Similarity (STS) and textual entailment.  Semantic Textual Similarity has been the focus of several SemEval tasks staring in 2012 \citep{agirre-EtAl:2012:STARSEM-SEMEVAL} and ongoing at the time of writing this paper\footnote{\url{http://ixa2.si.ehu.eus/stswiki}}. Given a pair of sentences, s1 and s2, STS systems compute how similar s1 and s2 are and return a similarity score bounded by the grade being used. 
STS is related to both paraphrasing and textual entailment, but instead of being binary it reflects a graded notion. It also differs from textual entailment in that it is not directional. STS is an enabling technology with application in Machine Translation evaluation, Information Extraction, Question Answering and Text Summarization.
Our work reuses existing STS datasets, and adds an interpretable layer, in the form of typed alignments between sentence segments.

Our formalization of the interpretable layers proposes the explicit alignment of segments, where each alignment is labeled with a relation type and a similarity score.
Previous work on alignment between text segments in the same language\footnote{As opposed to alignment of parallel corpora in machine translation settings.} have usually focused on the word level, 
with some exceptions. 
For instance, \citet{brockett2007aligning} released the 2006 PASCAL corpora composed of sentence pairs, where semantically equivalent words and phrases in the Text (T) and Hypothesis (H) sentences were aligned. Each word of H was either linked to one or more words of T or it was left unlinked, and the links were marked as either sure or possible depending on the degree of confidence in the alignment.  Annotators of the dataset viewed the sentence pairs of the corpora as pairs of parallel strings of words with lines of association between them, with limited coverage of some phrases like multiword expressions. In our work we go one step further and focus on text segments beyond words, as well as adding alignment types and similarity scores. 
In a similar effort, \citet{rus2012similar} 
aligned tokens from a STS dataset, although some short phrases were also aligned, such as chunks which were semantically equivalent but non-compositional. In our case our formalization covers all kind of segments, including non-equivalent and equivalent segments, compositional or not. 

In recent work, which has been performed in parallel to ours, 
\cite{pavlick-EtAl:2015:ACL-IJCNLP} annotated an automatically derived database of paraphrases for short phrases \citep{ganitkevitch2013ppdb} with entailment relations from Natural Logic \citep{maccartney-manning:2008:PAPERS}. They used crowdsourcing to annotate by hand around 14 thousand phrase pairs in the database. Section \ref{formalization} includes a head-to-head comparison of the annotation schemes, showing that this work is complementary to formalization and annotation.

In a different strand of work coming from the educational domain and close to textual entailment, \citet{nielsen2009recognizing} defined so-called facets, where each facet was a pair of words and a non-explicit semantic relation between both words. Each facet in the hypothesis text, usually a sentence, is annotated with information of whether it is entailed by the reference text. In the context of tutoring systems, their dataset comprises student responses and reference answers.  
Each reference answer was decomposed by hand into its constituent facets. The student answers are annotated with a label for entailed facets of the corresponding reference answer, but, contrary to our proposal, there is no explicit alignment between facets, and the facets do not necessarily correspond with text segments, but rather represent pairs of words having an unknown semantic relation in the text. 
Our initial motivation for interpretable STS was similar to that of \citet{nielsen2009recognizing}, as we think interpretability is especially useful in the field of tutoring systems, but we depart from that work in explicitly aligning segments in both sentences, as well as providing labels for the relation and similarity scores. 

The idea of facets was later followed by \citet{levy2013recognizing}, which call it partial textual entailment. 
This approach is complementary to ours, in that they could also try to align facets and characterize the semantic relations as well as the alignment relations. From another perspective, the same way they enrich textual entailment datasets with partial entailment annotations, we also enrich STS datasets with explicit alignments, where our types are related to entailment relations. We will get back to this relation in Section \ref{formalization}, after we present our formalization in full.

Other related work includes a SemEval task related to tutoring systems that automatically score student answers, the Joint Student
Response Analysis and 8th Recognizing Textual Entailment
Challenge \citep{Dzikovska-EtAl:2013_STARSEM}.
This task is the first large-scale and non-commercial automatic short answer grading  competition \citep{burrows2015eras}. The goal of the mentioned task was to assess student responses to questions in the science domain, focusing on the correctness and completeness of the response content. In a typical scenario, they expected that a correct student answer would entail the reference answer. The goal of the mentioned task was to label the student answers according to different categories (i.e. correct, partially correct or incomplete, contradictory, irrelevant and out-of-domain). 
The task includes a pilot subtask where participants had to annotate facets. 
In our opinion, effective feedback needs to 
identify the specific text segments of the student answers that differ from the reference answer, which we do via alignments.

The dataset presented here has been previously used in a subtask of the STS task in SemEval 2015 \citep{agirre-EtAl:2015:SemEval}.\footnote{\url{http://alt.qcri.org/semeval2015/task2/index.php?id=proba}}
And, the system described in this article is an improvement of a system which participated in the task, as described in \cite{agirre:2015:SemEval}.


\section{Dataset description}
\label{TaskDescription}

This section presents the Interpretable STS dataset. We first introduce the  annotation procedure, followed by the source of the sentence pairs, the evaluation method, and inter-tagger annotation data.

\subsection{Annotation procedure}
\label{sec:annotation-procedure}

This section introduces the annotation (further details can be consulted in the annotation guidelines). \footnote{\url{http://alt.qcri.org/semeval2015/task2/data/uploads/annotation_guidelines_semeval-2015_task2_interpretablests.pdf}} Given a pair of sentences, this is the procedure to be followed by annotators:

\begin{enumerate}
\item First of all, the annotator identifies the segments in each sentence separately, regardless of the corresponding sentence in the pair.
\item Secondly, the annotator aligns the chunks in order, from the clearest and strongest correspondences to the most unclear or weakest ones.
\item Third, for each alignment, the annotator provides a similarity score.
\item Finally, the annotator chooses the label or tag for each alignment.
\end{enumerate}

\paragraph{Text segments} Segments are annotated according to the definition of \emph{chunks} \citep{abney1991parsing}: “a non-recursive core of an intra-clausal constituent, extending from its beginning to its head. A typical chunk consists of a content word surrounded by a constellation of function words, matching a fixed template”. 
When marking the chunks of each sentence, the annotator follows the CONLL 2000 task guidelines\footnote{\url{http://www.clips.ua.ac.be/conll2000/chunking/}}, which were adapted slightly for our purpose: 
The main clause is split in smaller chunks consisting on noun phrases, verb chains, prepositional phrases, adverbs and other expressions. Figure \ref{interpretable:chunks:adaptation} shows some examples of chunks.

In order to help the annotators, we run the sentences through a publicly available open-source chunker\footnote{\url{https://github.com/ixa-ehu/ixa-pipe-chunk}} trained on CONLL 2000 corpora \citep{agerri2014ixa}.

\begin{figure}[t]
\begin{lstlisting}[frame=single, numbers=none]
Noun phrases: [The girl] / [Bradley Cooper and JJ Abrams]
Verb chains: [is arriving] / [does not like]
Prepositional phrases: [at a time] / [with the telescope] / [the house] [of that man]
Adverbial phrases: [of course]
Other expressions: [once upon a time] / [by the way]
\end{lstlisting}
\caption{Examples of chunks.}
\label{interpretable:chunks:adaptation}
\end{figure}

\paragraph{Alignment} The alignment is marked using an interface\footnote{We modified a tool developed by LDC to align words \url{https://www.ldc.upenn.edu/language-resources/tools/ldc-word-aligner}. We reused their XML-based annotation format as well. }. 
When aligning, the meaning of the chunks in context are taken into account. Annotators must try to align as many chunks as possible. Given some limitations in the interface, we decided to focus on one-to-one alignments, that is, one chunk can be aligned with at most one chunk. For this reason,
when having two options to align, only the strongest corresponding chunk will be aligned. The other chunk(s) will be left {\it unaligned}, and labeled with \textit{ALIC}. Chunks can be  also left unaligned if no corresponding chunk is found (\textit{NOALI} label). 
Punctuation marks are ignored, and left unaligned. 

\paragraph{Score} Once chunks are aligned, the annotator provides a similarity  score for each alignment, where the score ranges from 5 (maximum similarity, equivalence) to 0 (no relation at all). Note that an aligned pair would never score 0, as that would mean that the two chunks should not be aligned. See below for further restrictions concerning possible score values for specific labels.

\begin{table}[t]
\centering
\begin{tabular}{l c c c }
 \hline
 Label 	   & Chunk1  & Chunk2 & Score   \\
\hline
EQUI 		   & abduct   & kidnapped  & 5   \\ 
OPPO 		   & soar     & slump   & 4 	 \\   
SPE1 		   & two mountain goats    & two animals   & 1     \\ 
SPE2 		   & in Pakistan    & in NW Pakistan  & 4    \\ 
SIMI 		   & Russia   & South Korea  & 3     \\ 
REL 		   & on the porch    & on a couch   & 2    \\ 
 
\hline
\end{tabular}
\caption{Examples of aligned chunks, with label and score. } 
\label{chunk-examples} 
\end{table}

\paragraph{Label} When assigning labels to aligned chunks, the interpretation of the whole sentence, including common sense inference, has to be taken into account. The possible labels are the following:

\begin{description}
\item{\it EQUI}, both chunks are semantically equivalent;
\item{\it OPPO}, the meanings of the chunks are in opposition to each other;
\item{\it SPE1 or SPE2}, chunk in sentence 1 is more specific than chunk in sentence 2 (or vice versa);
\item{\it SIMI}, the meaning of the chunks are similar, and the chunks are not {\it EQUI}, {\it OPPO}, {\it SPE1}, or {\it SPE2};
\item{\it REL}, the meaning of the chunks are related, but they are not {\it SIMI}, {\it EQUI}, {\it OPPO}, {\it SPE1}, or {\it SPE2};
\end{description}

These six labels are exclusive, and each alignment should have one and only one such label. Some examples are provided in Table \ref{chunk-examples}. In addition, the following optional labels can be used in any alignment:
\begin{description}
\item{\it FACT}, the factuality, i.e. whether the statement is or is not a fact or a speculation is different  in the aligned chunks. 
\item{\it POL}, the polarity, i.e. the expressed opinion (positive, negative or neutral) is different in the aligned chunks. 
\end{description}

Note that {\it ALIC} and {\it NOALI} can also be {\it FACT} or {\it POL}, even not aligned, meaning that the respective chunk adds a factuality or polarity nuance to the sentence.

Labels and scores are not independent. After annotating scores and labels, the annotator should see that the following constraints are enforced: 

\begin{itemize}
\item {\it NOALI} and {\it ALIC} should not have scores but the {\it 0} value.
\item {\it EQUI} should have a 5 score. 
\item The rest of the labels should have a score larger than 0 but lower than 5. 
\end{itemize}

\begin{figure*}[t]
 \begin{center}
   \includegraphics[scale=0.65]{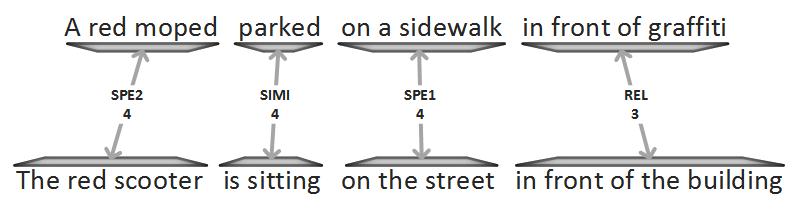}
 \end{center}
  \caption{Graphical representation of the interpretability layer. 
The two sentences were split in four segments each, which were aligned as follows: \emph{``A red moped''} is more general than \emph{``The red scooter''} with a similarity score of 4 (SPE2 4), \emph{``parked''} is similar to \emph{``is sitting''} with score 4 (SPE2 4), \emph{``on a sidewalk''} is more specific than \emph{``on the street''} with score 4 (SPE1 4), and \emph{``in front of graffiti''} is related to \emph{``in front of the building''} with score 3 (REL 3). See Section \ref{sec:annotation-procedure} for more details on the annotation procedure.}
  \label{example2}
\end{figure*}

\subsection{Source of the dataset}
\label{sec:source-dataset}

The dataset comprises pairs of sentences from news headlines (Headlines) and image descriptions (Images). We already mentioned a sample pair from Headlines (cf. Figure \ref{example}), and Figure \ref{example2} shows a sample pair from Images, together with their alignment. 
The Headlines corpus is composed of naturally occurring news headlines gathered by the Europe Media Monitor engine from several different news sources (from April 2nd, 2013 to July 28th, 2014) as described by \citet{Best05}.
The Images dataset is a subset of the PASCALVOC-2008 dataset, as described by \citet{Rashtchian:2010:CIA:1866696.1866717}, which consists of 1000 images with around 10 descriptions each.
The dataset comprised 756 and 750 sentence pairs from Headlines and Images, respectively, which were split evenly in  training and testing subsets. The dataset is freely available\footnote{\url{http://alt.qcri.org/semeval2015/task2/data/uploads/sts2015-interpretability-train.v3.tgz} and \url{http://alt.qcri.org/semeval2015/task2/data/uploads/test_evaluation_task2c.tgz}}. Table \ref{headlines_statistics} describes the statistics for the Headlines and Images datasets.

In addition, table \ref{headlines_statistics} shows some statistics for each of the datasets. Headlines contain slightly less chunks and less tokens per chunk than image captions. More than half of the aligned pairs in both datasets have a score of 5 (which corresponds to EQUI pairs)
with a decreasing number of aligned pairs for each score range. Regarding the labels, EQUI is the most used label, followed by SIMI, SPE1 and SPE2, REL and OPPO. The breakdown in scores and types is very similar in both datasets. ALIC is used a few times, more often in the Headlines dataset. There is a large number of unaligned chunks, specially in the Images dataset. Finally, FACT and POL are seldom used in the news dataset, and never in the Images dataset.

\begin{table}[t]
\centering
\begin{tabular}{l r r r r | r r r r}
\hline
  & \multicolumn{4}{c}{Headlines}\vline & \multicolumn{4}{c}{Images} \\
                   & Train & Test & All &\%   & Train & Test & All &\% \\
\hline
 Sentence pairs  	   & 378   & 378  & 756	 & & 375   & 375  & 750 & \\ 
 Chunks/sentence   & 4.2     & 4.2   & 4.2  & & 4.5   & 4.5  & 4.5 &  \\
 Tokens/chunk 	   & 1.9   & 1.9 & 1.9      & & 2.2     & 2.3    & 2.25   & \\ 
 \hline
 Aligned pairs 	   & 1064  & 1102 & 2166 & & 969   & 942  & 1911 &\\
\hline
 \hspace{0.2cm} Score $\in$ [5]   & 652   & 665  & 1317 &60.8\% & 529   & 499  & 1028 &53.8\%\\
 \hspace{0.2cm} Score $\in$ [4,5) & 189   & 225  & 414 &19.1\%  & 247   & 268  & 515  &26.9\%\\
 \hspace{0.2cm} Score $\in$ [3,4) & 133   & 126  & 259  &12\% & 101   & 107  & 208 &10.9\% \\
 \hspace{0.2cm} Score $\in$ [2,3) & 80    & 70   & 150	&6.9\%  & 75    & 55   & 130 &6.8\% \\
 \hspace{0.2cm} Score $\in$ [1,2) & 10    & 16   & 26 	&1.2\%  & 17    & 13   & 30  &1.6\% \\
\hline
 \hspace{0.2cm} EQUI 		   & 652   & 665  & 1317 &60.8\% & 529   & 499  & 1028 &53.8\%\\ 
 \hspace{0.2cm} SPE1 		   & 98    & 99   & 197   &9.1\% & 108   & 126  & 234  &12.3\%\\ 
 \hspace{0.2cm} SPE2 		   & 86    & 108  & 194   &8.9\% & 129   & 109  & 238 &12.4\% \\ 
 \hspace{0.2cm} SIMI 		   & 171   & 154  & 325   &15.0\% & 174   & 170  & 344 &18\% \\ 
 \hspace{0.2cm} REL 		   & 48    & 66   & 114   &5.3\% & 29    & 35   & 64  &3.3\% \\ 
 \hspace{0.2cm} OPPO 		   & 9     & 10   & 19 	  &0.9\% & 0     & 3    & 3  &0.2\%  \\ 
\hline
 ALIC		   & 92    & 99   & 191   & & 53    & 39   & 92  & \\ 
 NOALI 		   & 949   & 841  & 1790  & & 1406  & 1468 & 2874 & \\ 
 FACT	   & 10    & 20   & 30 	  & & 0     & 0    & 0   & \\
 POL 	   & 3     & 0    & 3	  & & 0     & 0    & 0  &  \\
\hline
\end{tabular}
\caption{Headlines and Images dataset statistics across splits. The first three rows report, respectively, the number of sentence pairs, chunks per sentence and tokens per chunk. The rows below report the number of aligned chunk pairs, with a break-down according to the similarity score, followed by a breakdown according to the label of aligned pairs. The last four rows report the number of unaligned chunks (ALIC, NOALI), and how many times the additional FACT and POL labels are used. } 
\label{headlines_statistics} 
\end{table}

\subsection{Annotation Effort}

The annotation of the 1501 pairs took 70 hours, 2.8 minutes per pair. The annotation was faster towards the end of the project, at around 2.3 minutes per pair. We used an in-house adaptation of an interface designed for cross-lingual word alignment\footnote{\url{https://www.ldc.upenn.edu/language-resources/tools/ldc-word-aligner}}, which helped to enter the annotations faster.

\subsection{Evaluation measures}
\label{sec:evaluation-measures}

In order to evaluate systems which perform Interpretable STS, we decided to adopt word alignment evaluation methods from the Machine Translation community. In particular, the evaluation method is based on 
that of  \citet{melamed1998manual}, which uses the F1 of precision and recall of token alignments. Note that \citet{fraser2007measuring} argued that F1 is a better measure than Alignment Error Rate.
The idea is that segment alignment is mapped into token alignment, where all token pairs in the aligned pairs are aligned with some weight.  The weight of each token-token alignment is the inverse of the number of alignments of each token, the so-called fan out factor \citep{melamed1998manual}. Precision is measured as the ratio of token-token alignments that exist in both system and gold standard files, divided by the number of alignments in the system. Recall is measured similarly, as the ratio of token-token alignments that exist in both system and gold-standard, divided by the number of alignments in the gold standard. Precision and recall are evaluated for all alignments of all pairs in one go.

The evaluation is done at four difference levels: segment alignment alone (ignoring labels and scores), segment alignment where we require that labels agree (i.e. pairs of segments with different labels are ignored), segment alignment where differences in score are penalized, and finally, segment alignment score where both labels and scores are taken into account. The evaluation script is freely available together with the dataset.

\subsection{Quality of annotation}
\label{QualityAnnotation}

To measure the viability and quality of the annotation and to calculate the {\it inter-tagger agreement} (ITA), two annotators annotated, individually, a random subset of 20 sentence pairs, 10 from each dataset.
Both annotators previously read the guidelines and agreed on mutual uncertainties. The agreement was computed using the evaluation script, where one tagger was taken as the system and the other one as the gold standard. Overall results for the agreement are shown in table \ref{ITA_results_detailed}.
Notice that the metrics used to compute the ITA are the same as the metrics used to evaluate system performance on the task. 

The segment alignment is done with very high agreement, both for Headlines and Images dataset. The agreement on the type is also high, in the 80s, as well as the agreement on scores (over 80). When considering the agreement on both type and score, the scores are also over 70, with the highest score for the simpler Images dataset. The high results show that the annotation task is well-defined and replicable, with high agreement scores.

\begin{table}[t]
\centering
\begin{tabular}{lll}
\hline
  &  Headlines  & Images\\
\hline
 Alignment     &     0.9139 &  1.0000 \\
 Type          &     0.7708 &  0.8095 \\
 Score         &     0.8473 &  0.9524 \\
 Type + Score  &     0.7375 &  0.7714 \\           
\hline
\end{tabular}
\caption{ Inter-tagger agreement on the Headlines and Images dataset.}
\label{ITA_results_detailed} 
\end{table}


\section{Relation to other annotation schemes}
\label{formalization}

\begin{table}[t]
\centering
\begin{tabular}{lll}
\hline
iSTS & NL          \\ \hline
EQUI & $\equiv$    \\
OPPO & $\neg$      \\
SPE1 & $\sqsubset$ \\
SPE2 & $\sqsupset$ \\
SIMI & $\sim$      \\
REL  & $\sim$      \\ \hline
\end{tabular}
\caption{Relation between our alignment types and the Natural Logic entailment relations used by \cite{pavlick-EtAl:2015:ACL-IJCNLP}}
\label{tab:formalization}
\end{table}

Our labels are closely related to those used in Natural Logic
\citep{maccartney-manning:2008:PAPERS}, and later adapted for the
purpose of annotating a database of paraphrases
\citep{pavlick-EtAl:2015:ACL-IJCNLP}. 
We compare our annotations to the
latter, as it is closer to this work.
They use a set of six mutually exclusive entailment relations:
\begin{itemize}
\item Equivalence (couch $\equiv$ sofa)
\item Opposites (old $\neg$ young)  
\item Forward/backward entailment (crow $\sqsubset$  bird)
\item Related by something other than entailment (boy $\sim$ little) 
\item Unrelated (professor $\#$ cucumber) 
\end{itemize}

Our labels have been created independently from theirs, but the overlap between both annotation schemes is remarkable. Table
\ref{tab:formalization} shows the mapping between the respective labels, which is a one-to-one mapping with exception of their $\sim$, where we
distinguish between similar (SIMI) and related
(REL). Several researchers have argued about the convenience to separate these phenomena \citep{Agirre:2009:SSR:1620754.1620758,hill2015cl}, where similarity refers to conceptually similar concepts (e.g. Russia - Korea) and relatedness refers to concepts which are closely related (e.g. Russia - Putin). Note that relatedness also encompasses similarity, as similar concepts also tend to be related. \citet{hill2015cl} present a recent review of these two phenomena, and a review of word relatedness and word similarity datasets. 

Our annotated resource is thus complementary to that of
\cite{pavlick-EtAl:2015:ACL-IJCNLP}, who have annotated via
crowdsourcing a subset of the 14K phrase pairs of an automatically
derived paraphrase database \citep{ganitkevitch2013ppdb}. In our case,
we  annotate 4K pairs\footnote{Note that, contrary to the other work, the 4K exclude pairs with no relation. One could easily derive non-related chunk pairs pairing all combinations of non-aligned chunks. }
of manually identified chunks in text pairs. Note that the source of pairs is different: while they label pairs of phrases which have been automatically induced as being paraphrastic, we label pairs of chunks which occur in pairs of naturally occurring sentences from different similarity ranges. In addition, we distinguish between similar and related pairs, and label explicitly factuality and polarity phenomena.


\section{System construction}
\label{SystemConstruction}

A system for Interpretable STS needs to perform chunking, align the
chunks, label and score the alignments. We
first describe a baseline system which performs each of the steps in turn, and then present improvements.

\subsection{Baseline construction}
\label{sec:basel-constr}

The baseline performs each of the steps following some publicly available algorithms. It first 
runs the ixa-pipes chunker\footnote{\url{https://github.com/ixa-ehu/ixa-pipe-chunk}} \citep{agerri2014ixa}.
We then lower-case all tokens and align identical tokens. Chunks are aligned based on the number of aligned tokens in a greedy manner, starting with the pair of chunks with the highest relative\footnote{The mean number of tokens is used for normalization.} number of aligned tokens.  
Chunks with no aligned tokens are left unaligned. 
Finally, the baseline uses a rule-based algorithm to directly assign labels and scores, as follows:
aligned chunk pairs are assigned the {\it EQUI} label, and the rest are either assigned {\it ALIC} (if they contain aligned tokens), or  {\it NOALI} (if they do not contain aligned tokens). The procedure to assign scores follows the alignment guidelines: {\it EQUI} pairs are scored with the maximum score and the rest are  scored with {\it 0}.

\subsubsection{Chunking}
\label{systemmodule1}

Given that the chunker is not perfect, we analyzed the output of the chunker with respect to the gold chunks available in the training data, and used some regular expressions to improve the chunks. The rules concern  conjunctions, punctuation and prepositions, where the rules are used to join adjacent chunks. 
The rules mainly join preposition and noun phrase into a single chunk, as well as noun phrases separated by punctuation or conjunctions, or a combination of those.
In addition to the chunker, we run the Stanford NLP parser \citep{klein2003accurate}, producing part of speech, lemma  and dependency analysis.

\subsubsection{Alignment}
\label{systemmodule2}

We use a freely available state-of-the-art monolingual word aligner \citep{sultan2014back} for producing token alignments. 
In order to produce the chunk alignment, each possible chunk alignment is weighted according to the number of aligned tokens in the chunks. 
The Hungarian-Munkres algorithm \citep{munkres1957algorithms} is then used to find the chunk alignments which optimize the overall alignment weight. 

\subsubsection{Labeling}
\label{systemmodule3}

\begin{table}[t]
\centering
\begin{tabular}{ll}
\hline
 \# & Feature description \\
\hline
 1 & Jaccard overlap \\
 2 & Jaccard overlap of non stopwords \\
 3 & Jaccard overlap of stopwords \\
 4 & Difference in length betweeen chunks 1 and 2 \\
 5 & Difference in length betweeen chunks 2 and 1 \\ 
 6 & Max WordNet path similarity among sense pairs \citep{wn_path} \\ 
 7 & Max WordNet LCH similarity among sense pairs \citep{wn_lch} \\ 
 8 & Max WordNet JCN similarity among sense pairs \citep{wn_jcn} \\ 
 9 & Same as 6 but simulating root with the maximum common subsumer \\ 
 10 & Same as 7 but simulating root with the maximum common subsumer \\ 
 11 & Same as 8 but simulating root with the maximum common subsumer \\
 12 & Whether chunk 1 senses are more specific than chunk 2 senses \\ 
    & in the WordNet hierarchy \citep{miller1995wordnet} \\
 13 & Whether chunk 2 senses are more specific than chunk 1 senses \\
    & in the WordNet hierarchy\\
 14 & Difference in WordNet depth of segment head\\
 15 & Minimum value of pairwise difference of WordNet depth\\
 16 & Maximum value of pairwise difference of WordNet depth \\
 17 & Lemmatized lowercased tokens of chunk 1 \\
 18 & Lemmatized lowercased tokens of chunk 2 \\
 19 & Maximum similarity value using first resource in Section \ref{systemmodule4}\\
 20 & Maximum similarity value using second resource in Section \ref{systemmodule4}\\
 21 & Maximum similarity value using third resource in Section \ref{systemmodule4}\\
\hline
\end{tabular}
\caption{Features used by the supervised classifier to assign labels to aligned chunk pairs.}
\label{SVM_features} 
\end{table}

Alignments are labeled using a multiclass supervised classification algorithm, trained with positive alignments in the training data\footnote{We extracted all aligned pairs with \textit{EQUI}, \textit{OPPO}, \textit{SPE1}, \textit{SPE2}, \textit{SIMI} and \textit{REL} labels.}. We use twenty one features including
token overlap, chunk length, WordNet similarity between chunk heads and WordNet depth.
The features are listed in table \ref{SVM_features}.

We used Support Vector Machines \citep{chang2011libsvm}.  
As training data is reduced we indistinctly joined the available datasets and performed grid search to optimize the cost and gamma parameters using randomly shuffled 5-fold cross validation. In these development experiments we found that the classifier was failing to detect FACT and POL, so we removed these labels from the training in the final system.

The development experiments also showed that the performance of the classifier was sensitive to the quality of the chunker. The classifier was first trained and tested using cross-validation on data which contained gold chunks and gold alignments, but when we run the classifier on test folds which contained system chunking, the performance suffered. We tried several variations of gold and automated versions of the train data, and obtained the best cross-validation results using the following versions in the training folds: a concatenation of the gold version and a version mixing automatic chunking and gold-standard alignments and labels. We thus trained the final classifiers on this semi-automatically produced version.

\subsubsection{Scoring module}
\label{systemmodule4}

The scoring module uses a variety of word similarity resources, as follows:
\begin{enumerate}
\item Euclidean distance between Collobert and Weston Word Vector \citep{Collobert:2008:UAN:1390156.1390177}. The distances $d$ were converted to similarity $s$ in the $[0..1]$ range using the following formula $1-d/max(D)$ where $D$ contains all distances observed in the dataset. 
\item Euclidean distance between Mikolov Word Vectors \citep{DBLP:journals/corr/abs-1301-3781}. 
The distance was converted into similarity as above. 
\item PPDB Paraphrase database values \citep{ganitkevitch2013ppdb}. We used the XXXL version. This resource yields conditional probabilities. As our scores are undirected, in case the database contains values for both directions, we average.
\end{enumerate}

Given a pair of aligned chunks   ($C_1$ and $C_2$), we compute the similarity for any word pair $sim(w,v)$ in the chunks,  where $w \in C_1$ and $v \in C_2$, as the maximum of the similarities according to the three  resources above. We then compute the  similarity between the chunks as the mean of two similarities, the addition of similarities for each word in the first chunk and the addition of similarities for each word in the second chunk, as follows:

\begin{align*}
sim(C_{1},C_{2}) = \frac{1}{2} & \left( \frac{\sum_{w \in C_{1}} (max_{v \in C_{2}} sim(w,v) *idf(w))}{\sum_{w \in C_{1}} idf(w)} \right.\\
& + \left. \frac{\sum_{w \in C_{2}} (max_{v \in C_{1}} sim(w,v) *idf(w))}{\sum_{w \in C_{2}} idf(w)} \right)
\end{align*}

In the equation above $idf$ is the inverse document frequency, as estimated in using Wikipedia as a corpus.

\subsection{Three systems}
\label{sec:three-systems}

We developed  three systems: the baseline (\textsc{Base}, cf. Section \ref{sec:basel-constr}), an improved baseline (\textsc{Base+}) with better chunking and alignment models (cf. Sections \ref{systemmodule1} \ref{systemmodule2}) but baseline labeling and scoring modules, and the full system (\textsc{Full}) with  supervised labeling and  similarity-based scoring (Sections \ref{systemmodule3} and \ref{systemmodule4}). The systems were  developed using the training subset of the dataset alone (cf. Section \ref{TaskDescription}), with no access to the test.


\section{Evaluation}
\label{Evaluation}
This section explains the results of the three systems we have developed and an error analysis of them. Then, a comparison with respect to the state-of-the-art is presented. 

\subsection{Developed systems}
\begin{table*}[t]
\centering
\small
\begin{tabular}{lcccc|cccc}
\hline
& \multicolumn{4}{c|}{\bf Headlines} & \multicolumn{4}{c}{\bf Images}  \\
                           & ALI             & TYPE            & SCORE           & T+S             & ALI             & TYPE            & SCORE           & T+S             \\ \hline
\textsc{Base}              & 0.6701          & 0.4571          & 0.6066          & 0.4571          & 0.7060          & 0.3696          & 0.6092          & 0.3693          \\
\textsc{Base+}             & \textbf{0.7709} & {0.5019} & 0.6892          & {0.5019} & \textbf{0.8388} & 0.4450          & 0.7280          & 0.4447          \\ 
\textsc{Full}    & \textbf{0.7709}          & \textbf{0.5343}          & \textbf{0.7007}          & \textbf{0.5220}          & \textbf{0.8388}          & \textbf{0.6091}          & \textbf{0.7612}          & \textbf{0.5884}          \\ \hline
\textsc{Full$_{GChunks}$}    & 0.8991          & 0.6402          & 0.8211          & 0.6185          & 0.8846          & 0.6557          & 0.8085          & 0.6159          \\ \hline
\end{tabular}
\caption{ Results (F1) of our three systems on each of the datasets. Columns show the results on each evaluation criteria, where T+S stands for ``Type and Score''. Best results in bold. The last row shows the results for the \textsc{Full} system when using gold standard chunks instead of automatically produced chunks.}
\label{STS2c_syschunk_table} 
\end{table*}

We evaluated the three systems (\textsc{Base}, \textsc{Base+} and \textsc{Full}) according to the evaluation measures set in Section \ref{sec:evaluation-measures}. 
Table \ref{STS2c_syschunk_table} shows the results on the Headlines and Images datasets.
The better chunking and alignment (\textsc{Base+} and \textsc{Full}) improves the alignment F1 score more than 10 points in both datasets with respect to \textsc{Base}. The poor performance in  alignment causes the baseline system to also attain low F1 scores in type and score, as well as the overall F1 score (T+S). The comparison between  \textsc{Base+} and \textsc{Full} shows that the classifier is able to better assign types, specially for Images. The method to produce the score is also stronger in \textsc{Full}, and thus produces the best overall F1 (T+S). Note that the performance for the four available metrics decreases, as all metrics are bounded by ALI, and T+S is bounded by both TYPE and SCORE.

All in all, the alignment results are strong, but the decrease of performance when taking into account the type shows that this is the most difficult task right now, with score being an easier task. In fact, had the labeling been perfect, the TYPE F1 score would be the same as ALI F1 score, but a drop around 23 absolute points is observed in both datasets. Regarding the two datasets, Headlines are more challenging, with lower scores across the four evaluations. 
\subsection{Error analysis}
We performed an analysis of the errors performed by the \textsc{Full} system at each level of processing, starting with chunking. The last row in Table \ref{STS2c_syschunk_table} reports the results of the \textsc{Full} system when running on gold standard chunks. The results improve for both datasets, with very high alignment results, and show that chunking quality is key for good performance. The results when using gold chunks are comparable for the two datasets, which indicates that the difference in performance for Headlines and Images when running the system on raw data (first three rows in Table  \ref{STS2c_syschunk_table}) is caused by the automatic chunker. We can thus conclude that Headlines is more difficult to chunk than Images, which causes worse performance on this dataset. Some of the errors in chunking seem to be related to verbs, as shown in an example below, and could be caused by the particular syntactic structures used in news headlines, which are different from those expected by the automatic chunker. \\
* [ Three ] [ feared ] [ dead ] {\bf{[ after helicopter crashes ]}} [ into pub ]\\
\vspace{0.1cm}
\hspace{0.7cm}
[ Three feared dead ] {\bf{[ after helicopter ] [ crashes ]}} [ into pub ] \\

Regarding the quality of the alignments, we found that the aligner tended to miss some alignments because it did not have access to semantic relations between words (e.g. cows and horse below) or numbers (500 and 580 below). The following pairs include in bold chunks which should have been aligned by the system:\\

{\bf{Two cows}} graze in a field .\\
\vspace{0.1cm}
\hspace{0.4cm}
{\bf{A brown horse}} in a green field .\\

Bangladesh building disaster death toll passes {\bf{ 500}}\\
\vspace{0.1cm}
\hspace{0.4cm}
Bangladesh building collapse : death toll climbs to {\bf{ 580}}\\

\begin{figure*}[t]
 \begin{center}
   \includegraphics[scale=1.15]{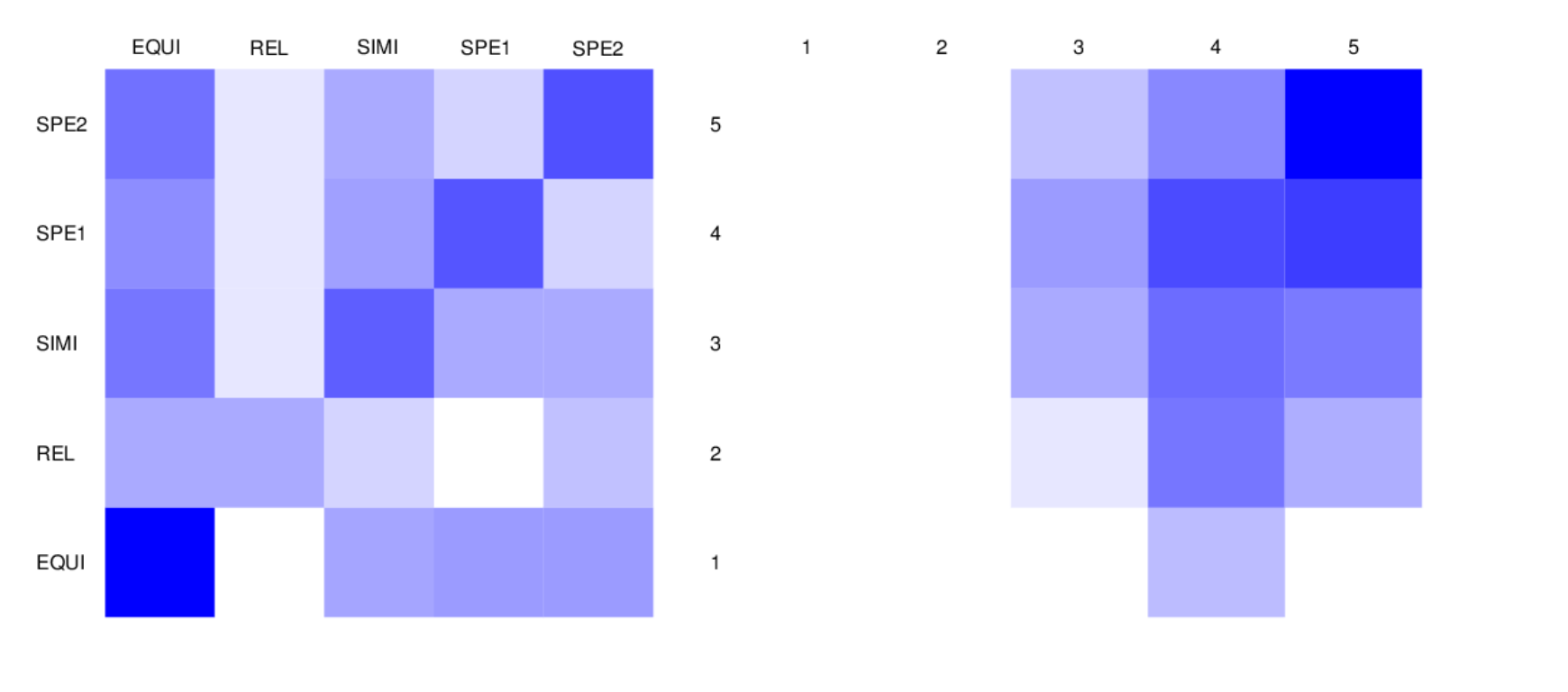}
 \end{center}
  \caption{Label and score confusion matrices as heatmaps (darker means higher counts in the cell) between the \textsc{Full} system and the gold standard scores and labels for the Headlines dataset.
Gold standard labels and scores in rows, system labels and scores in columns. }
  \label{confusion}
\end{figure*}

In order to check type-labeling errors, we built a confusion matrix between the \textsc{Full} system and the gold standard for the Headlines dataset (see Figure \ref{confusion}). The confusion matrix was built keeping correct alignments, as incorrectly aligned tokens cannot be analyzed for type errors. 
Most errors of the system are caused by the system being biased to return EQUI, which we think is caused by the imbalance of the classes in train (cf. Table \ref{headlines_statistics}). For example, in the next pair of sentences the aligned chunks {\bf{(in bold)}} should have been labeled as SPE2 instead of EQUI.\\

{\bf{A bus}} driving in a street .\\
 \vspace{0.1cm}
\hspace{0.4cm} 
{\bf{Red double decker bus}} driving down street .\\

In some cases, the system is not able to label equivalent chunks due to mistakes when recognizing identical entities or synonyms. In the next examples, the chunks of the sentence 2 {\bf{(in bold)}} have been labeled as more specific than the chunks of sentence 1 {\bf{(in bold)}}. However, in both sentence pairs the alignments should have been labeled as EQUI instead of SPE2.\\

Matt Smith {\bf{to leave}} Doctor Who after 4 years\\
\vspace{0.1cm}
\hspace{0.4cm}
Matt Smith {\bf{quits}} BBC 's Doctor Who

{\bf{De Blasio}} sworn in as New York mayor , succeeding {\bf{Bloomberg}}\\
 \vspace{0.1cm}
\hspace{0.4cm} 
{\bf{Bill De Blasio}} sworn in as New York mayor , succeeding {\bf{Michael Bloomberg}}\\

Regarding the errors in scores, Fig. \ref{confusion} shows the confusion matrix, where scores have been rounded to the nearest integer. Most errors are between contiguous scores, with some exceptions like the system returning 4 instead of 2, or 3 instead of 5. This shows a bias of our system towards high scores, which we would like to fix in the future. 

\subsection{Comparison to the state-of-the-art}
\begin{table*}[t]
\centering
\small
\begin{tabular}{lcccc|cccc}
\hline
& \multicolumn{4}{c|}{\bf Headlines} & \multicolumn{4}{c}{\bf Images}  \\
              & ALI             & TYPE            & SCORE           & T+S             & ALI             & TYPE            & SCORE           & T+S             \\ \hline
ExBThemis\_a  & 0.7032          & 0.4331          & 0.6224          & 0.4290          & 0.6966          & 0.3970          & 0.6068          & 0.3806          \\ 
ExBThemis\_m  & 0.7032          & 0.4331          & 0.6200          & 0.4288          & 0.6966          & 0.3970          & 0.6106          & 0.3870          \\ 
ExBThemis\_r  & 0.7032          & 0.4331          & 0.6209          & 0.4284          & 0.6966          & 0.3970          & 0.6092          & 0.3867          \\ 
RTM-DCU       & 0.4914          & 0.3712          & 0.4550          & 0.3712          & 0.3540          & 0.2283          & 0.3187          & 0.2282          \\ 
SimCompass\_c & 0.6467          & 0.4333          & 0.5636          & 0.3870          & 0.5433          & 0.2854          & 0.4545          & 0.2421          \\ 
SimCompass\_p & 0.6310          & 0.4284          & 0.5526          & 0.3872          & -               & -               & -               & -               \\ 
SimCompass\_w & 0.6461          & 0.4334          & 0.5619          & 0.3878          & 0.5428          & 0.2831          & 0.4561          & 0.2427          \\ 
UMDuluth\_1   & \textbf{0.7820} & 0.5058          & 0.6968          & 0.5004          & 0.8336          & 0.5529          & 0.7498          & 0.5431          \\ 
UMDuluth\_2   & \textbf{0.7820} & 0.5109          & 0.6986          & 0.5049          & 0.8336          & 0.5759          & 0.7511          & 0.5634          \\ 
UMDuluth\_3   & \textbf{0.7820} & {0.5154} & \textbf{0.7024}        & {0.5098}        & 0.8336          & 0.5605          & 0.7456          & 0.5473          \\ \hline
\textsc{Full}    & 0.7709          & \textbf{0.5343}          & 0.7007          & \textbf{0.5220}          & \textbf{0.8388}          & \textbf{0.6091}          & \textbf{0.7612}          & \textbf{0.5884}          \\ \hline
\end{tabular}
\caption{ Comparison to the state-of-the-art. Results (F1) on each of the datasets. Columns show the results on each evaluation criteria, where T+S stands for ``Type and Score''. Best results in each column in bold.}
\label{STS2c_syschunk_table_SOA} 
\end{table*}

 Table \ref{STS2c_syschunk_table_SOA} shows the results of our best system (\textsc{Full}) with respect to the state-of-the-art, as set in the SemEval Task 2 competition \citep{agirre-EtAl:2015:SemEval, karumuri2015umduluth:2015:SemEval, hanig2015exb:2015:SemEval,biccici2015rtm:2015:SemEval}, which included a subtask on Interpretable STS based on our dataset\footnote{Participants could send up to three runs. Note that the task also included a track where the gold chunks were made available to participants. For the sake of space, we focus on the most natural track, where systems need to chunk sentences on their own.}. Our system outperforms the best system (UMDuluth\_3) in both datasets, except in the ALI and SCORE results for Headlines. 
 
The UMDuluth\_3 system improved the quality of the publicly available OpenNLP chunker, with some post processing rules, which could explain the better performance of ALI on Headlines. They use the same alignment software as our system. The labeling module is a supervised system based on support vector machines, similar to ours. Our better results can be explained by a larger number of features, which include similarity scores from the scoring module and more WordNet similarity measures. Unlike ours, their scoring module is based on the labels. 

The good results of participating system and the improvement over baselines show that Interpretable STS is a feasible task in all steps: alignment, labeling of relations and scoring similarity. It is also indirect evidence that the task is well designed and the annotation consistent. 

Note that we participated in the Semeval task with a previous version of our system. The only difference is in the strategy to train the classifier for alignment labels, which was based on gold standard chunks and now uses a mixture of gold chunks and system-produced chunks  (cf. Section \ref{systemmodule3}).


\section{Application of Interpretable STS}
\label{ApplicationInterpretable}

In order to judge whether the information returned by an Interpretable STS system can be used to clarify and explain semantic judgments to humans, we performed two user studies. We first devised a verbalization algorithm, which, given two sentences, their similarity score and the typed and scored alignment between chunks, returns English text verbalizing the differences / commonalities between the two sentences. We then contrasted the activities of the users with and without the Interpretable STS verbalizations, trying to show that the verbalizations helped the users in the two case studies.

\subsection{Verbalization}
\label{sec:verbalization}

Given the output of the Interpretable STS system, we devised a simple template-based algorithm to verbalize the alignment information into natural language.
The label of the alignment is used to select which template to use, and the score is used to qualify the strength of the relation, as summarized in Table \ref{templates_verbalization}. 
An example of a verbalization for a sentence pair is shown in the bottom of Figure \ref{scenario4:example}.

\begin{table}[t]
\centering
\begin{tabular}{ll}
\hline
Label & Verbalization produced \\
\hline
 EQUI        &   \emph{X} and \emph{Y} mean the same  \\
 SPE1        &   \emph{X} is [a bit more $|$ more $|$ much more ] specific than \emph{Y}  \\
 SPE2        &   \emph{X} is [a bit more $|$ more $|$ much more ] general than \emph{Y}  \\
 SIMI        &   \emph{X} and \emph{Y} are [very $|$ $\emptyset$ $|$ slightly $|$ scarcely ] similar    \\
 REL         &   \emph{X} and \emph{Y} don't mean the same but are [closely $|$ $\emptyset$ $|$ somehow $|$ distantly] related  \\
 OPPO        &   \emph{X} and \emph{Y} mean the opposite  \\
\hline
\end{tabular}
\caption{Templates employed for producing verbalizations summarized by label. \emph{X} and \emph{Y} refer to the aligned chunks from sentence 1 and 2, respectively. The score is used to select the qualifiers in SPE1, SPE2, SIMI, and REL. }
\label{templates_verbalization}
\end{table}

We are aware that the verbalization algorithm could be improved, specially to avoid repetitions, and make the text more fluent and easier to read. It currently produces one sentence per alignment, resulting in too much text. The information from several alignments could be synthetized and summarized in shorter messages. In any case, we will show that this simple verbalization algorithm is effective enough in the two user case studies.  

\begin{figure}

\begin{lstlisting}[frame=single, numbers=none]
Please, evaluate the two sentences with a score between 0 and 5, with the following interpretation:

(5) The two sentences are completely equivalent, as they mean the same thing. 
      The bird is bathing in the sink. 
      Birdie is washing itself in the water basin.
(4) The two sentences are mostly equivalent, but some unimportant details differ.
      In May 2010, the troops attempted to invade Kabul.
      The US army invaded Kabul on May 7th last year, 2010.
(3) The two sentences are roughly equivalent, but some important information differs/missing.
      John said he is considered a witness but not a suspect. 
      "He is not a suspect anymore."
(2) The two sentences are not equivalent, but share some details.
      They flew out of the nest in groups. 
      They flew into the nest together.
(1) The two sentences are not equivalent, but are on the same topic.
      The woman is playing the violin.
      The young lady enjoys listening to the guitar.
(0) The two sentences are completely dissimilar.
      John went horse back riding at dawn with a whole group of friends.
      Sunrise at dawn is a magnificent view to take in if you wake up early enough for it. 

Please, note that you have some explanations below the sentences. Read them carefully and use them to assign your scores.

------------------------------------------------------------------------

Afghan legislators approve new election law 
Afghan president approves new election law

They are very similar.
Note that 'Afghan legislators' and 'Afghan president' don't mean the same but are closely related in this context.
Note also that 'approve' and 'approves' mean the same in this context.
Note also that 'new election law' and 'new election law' are very similar in this context.

[ Write answer ]

\end{lstlisting}
\caption{Instructions and task for users participating in the first user study. This example shows a 
verbalization based on system's alignments.}
\label{scenario4:example}
\end{figure}

\subsection{First user study: STS}
\label{experiment1}

In the first user study, the volunteers need  to score the
similarity of the two sentences. Figure \ref{scenario4:example} shows the instructions for the volunteers, which mimic those used to annotate STS datasets \citep{agirre-EtAl:2015:SemEval}. The Figure corresponds to the case where a verbalization is shown to the volunteer. We then measured the agreement of the volunteers with the gold standard STS score. In order to contrast whether the verbalizations had any impact in the performance of the users in the task, we run three scenarios: no verbalization, automatic verbalization based on the Interpretable STS gold standard, automatic verbalization based on the Interpretable STS system output.

\subsection{Second user study: English students}
\label{experiment2}

\begin{figure}[t!]

\begin{lstlisting}[frame=single, numbers=none]

Professor Smith asked his students to write headlines after reading some texts.
Then he graded students using his own headlines as reference.
The grades used by professor Smith are the following ones: Insufficient (0-4.9), good (5-6.9), above good (7-8.9), excellent (9-10).
Your task is to evaluate the grading done by professor Smith from 0 to 10, being 0 complete disagreement and 10 complete agreement.
The first headline is the reference headline of professor Smith, the second one the headline of the student.

Afghan legislators approve new election law 
Afghan president approves new election law 
Grade: good

[ Write answer ]
\end{lstlisting}
\caption{Instructions and task for users participating in the second user study. In this example, no verbalization is given to the user. }
\label{scenario1:example}
\end{figure}

In the
second user study, we consider an English as a Second Language education scenario, where the
volunteers played the role of an inspector who is overseeing the grades given
by a lecturer to a student. The student had to summarize a piece of news into a single
headline. We re-use the pairs of sentences in the Headlines dataset, together with their similarity score. The volunteer is given two sentences: the first one is the reference headline used by the professor to asses the student, and the second  headline is  produced by the student. The similarity score is used as the grade given to the student.

The task given to the user is thus to assess to what extent they agree with the grading. Users were given the following information: the reference headline of the professor, the headline done by the student, the grade given, and, optionally, the feedback in the form of the automatically produced verbalization. We collect the feedback (agreement level) in the form of an integer between 0 (complete disagreement) and 10 (complete agreement). Figure \ref{scenario1:example} shows the instructions and one example pair, alongside the grade.

We run three scenarios: no verbalization, automatic verbalization based on the Interpretable STS gold standard, and automatic verbalization based on the Interpretable STS system output.

\subsection{Setting the task}
\label{sec:setting-task}

To conduct the user studies, we randomly selected 48 sentence pairs
from the Headlines dataset (see section \ref{TaskDescription}). The sentence pairs are accompanied by a gold standard similarity score which ranges from 0 (no similarity) to 5 (equivalence), and we thus  sampled the 48 pairs uniformly according to the score. The same set of 48 pairs was used in the two user studies.

The first user study
 involved 4 native English speakers. For the second user study, which was related to an  English as a Second Language setting, we involved 4 non-native English speakers with a verified level C1 of English.

To test whether verbalizations are useful or not, we randomly split the 48 items in 4 item sets (A, B, C and D) and distributed them among participants (E1-E4) according to the sketch shown in table \ref{evaluation:sketch}.

The sketch helps organize which files are distributed without verbalizations, which ones are distributed with verbalizations based on gold standard annotations of the Semeval data, and which ones are distributed with verbalizations produced by the system described in section \ref{SystemConstruction} (using the system chunk input data). The sketch distributes items across users and verbalizations in a uniform way in order to reduce biases across users, verbalizations and item sets. 
The same sketch has been used to distribute the files for both scenarios.

Rows from table \ref{evaluation:sketch} show how each item set with a specific verbalization (No verb, GS verb, SYS verb) is assigned to each participant and in which order. 
For instance, user E4 will do E4\_1, E4\_2 and E4\_3 in order, that is, the user will first do items in the item set D with no verbalization, then the item set C with GS verbalization and finally the item set B with SYS verbalization. We always show the no verbalized item set first, followed by verbalized itemsets, which are offered in different orders.

\begin{table}[t]
\centering
\begin{tabular}{llll}
\hline
Item sets & No verb & GS verb & SYS verb \\
\hline
 A        & E1\_1   & E2\_2   & E3\_2    \\
 B        & E2\_1   & E3\_3   & E4\_3    \\
 C        & E3\_1   & E4\_2   & E1\_2    \\
 D        & E4\_1   & E1\_3   & E2\_3    \\
\hline
\end{tabular}
\caption{Sketch used to distribute item sets (A-D) among participants (E1-E4) with the three possible verbalizations option in the rows. The number after the underscore refers to the order of presentation to the user, e.g. E2\_2 is shown to user E2 after E2\_1 and before E2\_3.}
\label{evaluation:sketch}
\end{table}

\subsection{Results}
\label{sec:results}

\begin{table}[t]
\centering
\begin{tabular}{llll}
\hline
        			& No verb & SYS verb & GS verb \\
\hline
Pearson $r$ 	&    0.83      &      0.92   &   0.90         \\
Spearman $\rho$ 		&    0.83      &      0.92   &   0.91         \\
\hline
\end{tabular}
\caption{  First user study: Correlations for non verbalized items,  gold standard verbalized items and system verbalized items.}
\label{evaluation:overall_correlation}
\end{table}

To measure the results of the first user study, we use the correlation between the scores given by participants and the gold standard STS score. We follow the tradition on the open evaluation tasks on STS \citep{agirre-EtAl:2012:STARSEM-SEMEVAL,agirre-EtAl:2015:SemEval} and use Pearson coefficient correlation as the main measure, but also report Spearman rank correlation. 
Table \ref{evaluation:overall_correlation} shows the correlation  for non-verbalized pairs, gold standard verbalized pairs, and system verbalized pairs.
Both correlation measures seem to output very similar values, 
with higher correlation values for the verbalized scenarios, showing that the explanations are indeed  helpful in this task. The verbalizations obtained from the system output are comparable to those of the gold standard, showing that approximate performance might be enough for being helpful in this task. Even if the amount of data points is small, we performed significance tests between the verbalization options using Fisher's z-transformation for
relatedness \citep[equation 14.5.10]{press2002numerical}. The difference between system verbalization and no verbalizations is statistically significant for both Pearson and Spearman\footnote{p-values of 0.057}, but the p-values for gold standard verbalizations vs. no verbalizations are larger\footnote{p-values of 0.178 on Pearson and 0.107 on Spearman}. Finally, the difference between system and gold standard verbalizations is not statistically significant.  

In the second  user study the results correspond to  the agreement level in each scenario (%
cf. Table \ref{evaluation:overall_agreements}). The table reports both the mean agreement level (the average of the raw agreement level introduced by the user), and the binary agreement (how many times the user entered an agreement of 5 or larger). 
In this user study, the effect of system verbalizations is not as clear as in the previous case: the binary agreement is better (83\% vs. 77\%) but the mean agreement level is very similar (7.6 vs. 7.4 agreement). The automatic verbalizations produced using gold standard annotations do have a clear impact in the task (94\% vs. 77\% binary agreement, and 8.8 vs. 7.4 agreement level), as the users tend to agree more with the scores assigned by the lecturer. The difference between system verbalization and no verbalizations is not statistically significant in any case, but the difference between gold verbalizations and no verbalization is significant \footnote{p-values of 0.019 and 0.031 for the agreement level and binary agreement, respectively, using paired t-test}.

All in all, the results show that a simple method to produce
verbalizations based on Interpretable STS annotations are effective in
both user studies, as the users could accomplish better the task at
hand. This is a strong indication that our annotation task is
well-defined, and leads to verbalizations which are intelligible and
which help the users understand the semantic similarity of the target texts.  The results obtained by the 
Interpretable STS systems are promising, with a clear positive effect
in the first user study.

\begin{table}[t]
\centering
\begin{tabular}{llll}
\hline
        			& No verb & SYS verb & GS verb \\
\hline
Agreement level	&     7.4 & 7.6 & 8.8    \\
Binary agreement	&     77\% & 83\% & 94\%    \\
\hline
\end{tabular}
\caption{Second user study: Agreement level with grade $[0..10]$ with non verbalized items,  gold standard verbalized items and system verbalized items. } 
\label{evaluation:overall_agreements} 
\end{table}


\section{Conclusions and future work}
\label{ConclusionsFutureWork}

This paper presents Interpretable
Semantic Textual Similarity, where we formalize an interpretability layer on
top of STS. We describe a publicly available dataset of sentence pairs, where the
relations between segments in each sentence are labeled with a relation
type and a similarity score. The labels represent relations between
segments such as equivalence, opposition, specificity, similarity and
relatedness, together with factuality and polarity differences. The
Interpretable STS labels are closely related to those available in
Natural Logic or Textual Entailment, and, thus, our dataset is
complementary to resources such as those presented in \cite{pavlick-EtAl:2015:ACL-IJCNLP}.

We have also built a system for Interpretable STS, based on a
pipeline which first identifies the chunks in each input sentence, then
aligns the chunks between the two sentences, and finally uses a supervised system
to label the alignments and a mixture of several similarity measures to score the
alignments.  The good results and the improvement over baselines show
that Interpretable STS is a feasible task in all steps: alignment,
labeling of relations and scoring of similarity. It is also indirect
evidence that the task is well designed and the annotation consistent,
as also supported  by the high inter-annotator agreement.

Beyond the low-level annotation we also studied whether the annotations could be
useful in final applications. To do so, we constructed a simple verbalization algorithm,
which given two sentences and the Interpretable STS annotations,
produces a textual explanation of the differences/similarities between
the sentences. We then carried out two succesfull small-scale user studies, which show
evidence that users which had access to the explanations perform the
task better. We take this as a preliminary indication that automatically produced explanations are effective to
understand the texts.

In the near future, we would like to improve the performance of the
Interpretable STS system. The current system performs each step
independently (alignment, labeling and scoring of the chunk pair), but does not enforce consistency. For instance, it can
produce a weak relation type like REL and a strong similarity score
such as 4.5, or vice versa. In fact, the alignment score could feed the
typing, and the type of the alignment could be useful for assigning
the score. We are thus currently exploring joint algorithms which
would perform some of the steps together, using neural networks as in \citep{zhou2016learning}. The error analysis shows that  our system has a bias towards equivalence and high scores, which future versions of the system should try to remedy.

We would also like to improve our simple and naive verbalization algorithm, as the
effectiveness in real tasks also depends on producing natural-looking
text which is up to the point and does not contain superfluous
information. Finally, we plan to perform a more extensive user
study on a real task. Tutoring systems for English as a second
language look like a promising direction for building systems which
can automatically grade students and  produce
explanations of the grading.

\section{Acknowledgements}
Aitor Gonzalez-Agirre and Inigo Lopez-Gazpio are supported by doctoral grants from MINECO. The work described in this project has been partially funded by MINECO in projects MUSTER and TUNER (TIN 2015-65308-C5-1-R), as well as the Basque Government (IT344-10). We also want to thank the volunteers that participated in the user studies. 

\bibliographystyle{elsarticle-harv}
\bibliography{main}

\end{document}